# Comparison of Support Vector Machine and Back Propagation Neural Network in Evaluating the Enterprise Financial Distress

Ming-Chang Lee[1] and Chang To[2]

[1]Department of Information Management, Fooyin University ,Taiwan
ming_li@mail2000.com.tw
[2]Department of Information Management Shu-Te University, Taiwan
changt@stu.edu.tw

## ABSTRACT

*Recently, applying the novel data mining techniques for evaluating enterprise financial distress has received much research alternation. Support Vector Machine (SVM) and back propagation neural (BPN) network has been applied successfully in many areas with excellent generalization results, such as rule extraction, classification and evaluation. In this paper, a model based on SVM with Gaussian RBF kernel is proposed here for enterprise financial distress evaluation. BPN network is considered one of the simplest and are most general methods used for supervised training of multilayered neural network. The comparative results show that through the difference between the performance measures is marginal; SVM gives higher precision and lower error rates.*

## 1. INTRODUCTION

The financial crisis means that business enterprise loses the ability of payment, has no capability to play expired liability or expenses and appears asset value less than issued debt. Once the enterprise occur financial crisis, it will bring great loss for executive and financial institution. Financial distress prediction is of interest not only to managers but also to external stakeholders of company. Gestel et al. [10] observed corporate bankruptcy cause substantial losses to the business community and society as whole. For financial institutions, poor decision when granting credit to corporations in either losing potentially valuable clients or incurring substantial capital loss when the client subsequently defaults.

From the late 1980s, the machine learning techniques in the Artificial Intelligence (AI) area, such as Artificial Neural Network (ANN) were applied to financial distress prediction studies [1] [4] [5] [6] [13] [14] [15] [23]. Singh et al.,[24] used ANN to the fault tolerant behavior. Chakraborty and Sharma [2] used Radial Basis Function Neural Network (RBF), multiplayer Perception (MLP), Self-Organized Competition (SOC) and Support Vector Machines (SVM) to examine credit evaluation capability. Currently, there has been a considerable interest in ANN because enterprise index with credit risk exist non linear relationship. It confirms that the effect of credit risk prediction using ANN is good than using MDA [18][20][23]. But, a potential drawback of ANN methods is that they need man-made adjustability in calculated process. The calculated processes consume labor power and time.

In order to improve the performance of enterprise financial distress evaluation, many artificial intelligence methods such as neural network have been widely used. These methods are based on an empirical and have some disadvantages such as local optimal solution, low convergence rate, and epically poor generalization when the number of class samples is limited [21]. In 1990s, Support Vector Machine (SVM) was introduced to cope with the classifications and the small number of quadratic programming in SVM training is applied. Fan and Palaniswami [8] applied SVM to select the financial distress predictors. SVM was introduced by vapnik in the late 1960's on the foundation on statistical learning theory [19]. SVM is a set of related supervised learning methods used for classification and regression. O'Neill and Penm [12] use SVM algorithm testing Standard & Poor (S&P) publisher credit data, which provide credit safe information to an investor. SVM is very effective method for general purpose pattern recognition based on structural risk minimization principles.

This paper is organized as follows: Section 2 deals with the foundations of support vector machine. In this section we will consider the problem of linear and nonlinear decision function Section 3 deals with back-propagation neural network. In section 4, we have a procedure for Enterprise Financial Distress evaluation, and give an illustration. It help in understanding of this procedure, a demonstrative case is given to show the key stages involving the use of the introduced concepts. Section 5 is conclusion.

## 2. Overview of Support Vector Machine

SVM creates a line or a hyper-plane between two sets of data for classification. Input data X that fall one side of the hyper-plane , $(X^T \bullet W - b) > 0$, arc labeled as +1 and those that fall on the other side, $(X^T \bullet W - b) < 0$ , arc labeled as -1.

**Definition 1: hyper-plane P**
Let $\{X_i, y_i\} \in R^n$ be training data set, $y_i \in \{1, -1\}$, i = 1 ,2,… ,n. There exits hyper-plan

$$P = \{X \in R^n \mid X^T \cdot W + b = 0\} \quad (1)$$

The training data set satisfies the following con

$$X_i^T W + b \geq 1, \quad y_i = 1 \quad (2)$$

$$X_i^T W + b \leq -1, \quad y_i = -1$$

It can be written as

$$y_i(X_i^T W + b - 1) \geq 0$$

**Definition 2: hyper-plane** $p^+$ and $p^-$

Let $\{X_i, y_i\} \in R^n$ be training data set, $y_i \in \{1, -1\}$, $i = 1, 2, \ldots, n$.

$$p^+ = \{X \in R^n \mid X_i^T W + b = 1\} \quad (3)$$

$$p^- = \{X \in R^n \mid X_i^T W + b = -1\}$$

The optimization problem presented in the proceeding section is hard because it depends on the absolute value of |W|. The reason is that it is a non-convex optimization problem, which is known to be much more difficult to solve the convex optimization problems. It is possible to alter the equation by substituting $||W||$ with $1/2 ||W||^2$ without changing the solution. The SVM problem is denoted as a quadratic programming (QP) optimization problem.

$$\text{Min} \quad 1/2 ||W||^2,$$
$$\text{s. t.} \quad y_i (x_i^T \cdot W + b) - 1 \geq 0, \quad 1 \leq i \leq n \quad (4)$$

Introducing Lagrange multipliers $\alpha_i$ to solve this problem of convex optimization and making some substitutions, we arrive to the Wolfe dual of the optimization problem:

$$L(w,b) = \frac{1}{2} ||W||^2 - \sum_{i=1}^{n} \alpha_i y_i [(x_i^T W + b) - 1] \quad (5)$$

We find the value of $W$ and b in $L(w,b)$, and set

$$\frac{\partial L(w,b)}{\partial w} = 0, \quad \frac{\partial L(w,b)}{\partial b} = 0 \quad (6)$$

The solution in (6) is the following condition

$$w = \sum_{i=1}^{n} \alpha_i y_i X_i \quad (7)$$

$$\sum_{i=1}^{n} \alpha_i y_i = 0$$

We substitute (7) into (5) and the dual form of SVM is:

$$\max w(\alpha) = \sum_{i=1}^{n} \alpha_i - (\sum_{i,j=1}^{n} \alpha_i \alpha_j y_i y_j (x_i \cdot x_j))/2$$

$$\text{s. t. } \sum_{i=1}^{n} \alpha_i y_i = 0 \quad (8)$$

$$0 \leq \alpha_i \leq c, \ i = 1, 2, ..., n$$

The number of variable, $\alpha_i$ in the problem is equal to the number of data-cases, n. The training cases with $\alpha_i > 0$, representing active constraints on the position of the support hyper-plane are called support vector, and $X_i \in p^+ or\ p^-$.

This is convex quadratic programming problem, so there is global maximum. There are a number of optimization routines capable of solving this optimization problem. The optimization can be solved in O ($n^3$) time and in linear time in the number of attributes. (Compare this to neural networks that are trained in O (n) time).

The SVM problem satisfies the KKT condition.

The KKT condition is:

$$\frac{\partial L(w,b)}{\partial w_j} = w_j - \sum_{i=1}^{n} \alpha_i \alpha_j X_{ij}. \quad j = 1, 2, ..., n$$

$$\frac{\partial L(w,b)}{\partial b} = -\sum_{i=1}^{n} \alpha_i y_i = 0 \quad (9)$$

$$y_i(w^T x_i - b) - 1 \geq 0$$

$$\alpha_i \geq 0$$

$$\alpha_i [y_i(w^T x_i - b) - 1] = 0$$

What we are really interested in the function f (.) which can be used to classify future test cases,

$$f(x) = w^{*T} x - b^* = \sum_i \alpha_i^* y_i x_i^T x - b^* \quad (10)$$

As an application of the KKT conditions we derive a solution for $b^*$ by using the complementary slackness condition,

$$b^* = (\sum_j a_j y_j x_j^T x_i - y_i) \ i\ a\ support\ vector \quad (11)$$

The hyper-plane decision function can thus be written as:

$$f(x) = sign(\sum_{i=1}^{n} \alpha_i^* y_i (x \cdot x_i) - b^*) \quad (12)$$

Where x is the input data to be classified and sign (.) is a signum function that outputs either +1 or -1 depending on the sign of the computed value inside the parentheses.

For problems that have a non-linear decision hyper-plane, a mapping function φ (x) is used to transform the original input space $R^n$ into a higher dimension Euclidean space $R^n$(H).

$$φ (x): R^n \longrightarrow R^n (H)$$

International Journal of Artificial Intelligence & Applications (IJAIA), Vol.1, No.3, July 2010

In this new space the optimal hyper-plane is derived. Nevertheless, by using kernel functions which satisfy the Mercer theorem, it is possible to make all the necessary operations in the input space by using $K(x, x_j) = \varphi(x) \cdot \varphi(x_j)$. That computes the dot product in the higher dimensional space, called a kernel function, is used both in training and in classification.

The decision function is formulated in terms of these kernels:

$$f(x) = sign(\sum_{i=1}^{n} \alpha_i^* y_i (\phi(X) \phi(X_i) - b^*))$$

$$= sign(\sum_{i=1}^{n} \alpha_i^* y_i (K(X, X_i) - b^*)) \qquad (13)$$

The $\alpha_i^*$, $b^*$ can be calculated by considering KKT condition [7].

Common examples of kernel function are: non-linear function [16]:

$K(x, x_i) = x^T x$; (Polynomial function) (14)

$K(x, x_i) = (x^T x + 1)^d, \quad d > 0;$ (15)

$K(x, x_i) = \tanh(x^T x' + 1)$ (Sigmoid function) (16)

$$K(x, x_i) = \exp(-\gamma \|x - x_i\|^2) \quad \text{(Radial Basis Function)} \qquad (17)$$

Where d is the degree of Polynomial kernel. When using some of the available SVM software packages or toolboxes a user should choose (1) Kernel function (e.g. Gaussian kernel) and its parameters, (2) constant C related to the slack variables. Several choices should be examined using validation set in order to find the best SVM.

**SVM program procedure**

Step 1: Transform data to the format of an SVM package
Step 2: Conduct simple scaling on the data
Step 3: Consider the RBF kernel
Step 4: Use cross-validation to find the best parameter C and γ [11].
Step 5: Use the best parameter C and γ to train the whole training set
Step 6: Test

## 3. BACK –PROPAGATION NEURAL NETWORK

The back propagation neural is a multilayered, feed forward neural network and is by far the most extensively used. Back Propagation works by approximating the non-linear relationship between the input and the output by adjusting the weight values internally. A supervised learning algorithm of back propagation is utilized to establish the neural network modeling. A normal back-propagation neural (BPN) model consists of an input layer, one or more hidden layers, and output layer. There are two parameters including learning rate (0 < α <1) and momentum (0 < η <1) required to define by user. The theoretical results showed that one hidden layer is sufficient for a BP network to approximate any continuous mapping from the input patterns to the output patterns to an arbitrary degree freedom [9]. The selection and nodes of hidden layers primarily affect the classification performance. The following figure shows the topology of the black-propagation neural network that includes and input layer, one hidden layer and output layer.

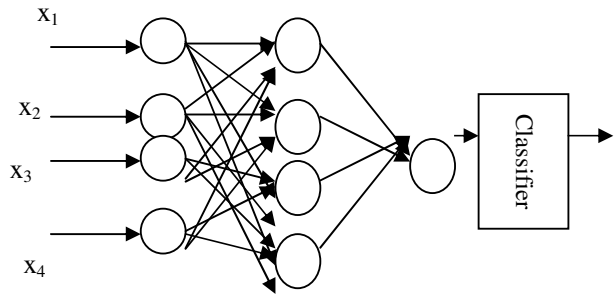

Figure 1. Two category model of back-propagation

**BPN program-Training process:**

Step 1: Design the structure of neural network and input parameters of the network.

Step2: Get initial weights W and initial θ (threshold values) from randomizing.

Step 3: Input training data matrix X and output matrix T.

Step 4: Compute the output vector of each neural units.

(a) Compute the output vector H of the hidden layer

$$net_k = \sum w_{ik} x_i - \theta_k \qquad (18)$$

$$H_k = f(net_k) \qquad (19)$$

(b) Compute the output vector Y of the output layer

$$net_j = \sum w_{kj} H_i - \theta_j \qquad (20)$$

$$Y_j = f(net_j) \qquad (21)$$

(c) Compute the root of mean square

$$RMS = \sqrt{\frac{\sum (y_j - T_j)^2}{n}} \qquad (22)$$

Step 5: Compute the distance δ

   (a) Compute the distance δ of the output layer

$$\delta_j = (T_j - Y_j) f'(net_j) \quad (23)$$

   (b) Compute the distance δ of the hidden layer

$$\delta_k = (\sum_j \delta_j w_{kj}) f'(net_j) \quad (24)$$

Step 6: Compute the modification of W and θ (η is the learning rate, α is the momentum coefficient)

   (a) Compute the modification of W and θ of the output layer

$$\Delta w_{kj}(n) = \eta \delta_j H_k + \alpha \Delta w_{kj}(n-1) \quad (25)$$

$$\Delta \theta_j(n) = -\eta \delta_j + \alpha \Delta \theta_j(n-1) \quad (26)$$

   (b) Compute the modification of W and θ of the hidden layer

$$\Delta w_{ik}(n) = \eta \delta_k X_i + \alpha \Delta w_{ik}(n-1) \quad (27)$$

$$\Delta \theta_k(n) = -\eta \delta_k + \alpha \Delta \theta_k(n-1) \quad (28)$$

Step 7: Renew W and θ

   (a) Renew W and θ of the output layer

$$w_{kj}(p) = w_{kj}(p-1) + \Delta w_{kj} \quad (29)$$
$$\theta_j(p) = \theta_j(p-1) + \Delta \theta_j \quad (30)$$

   (b) Renew W and θ of the hidden layer

$$w_{ik}(p) = w_{ik}(p-1) + \Delta w_{ik} \quad (31)$$
$$\theta_k(p) = \theta_k(p-1) + \Delta \theta_k \quad (32)$$

Step 8: Repeat step 3 to step 7 until converge.

**BPN program-testing process:**

Step 1: Input the parameters of the network

Step 2: Input the W and θ

Step 3: Input the unknown data matrix X

Step 4: Compute the output vector

   (a) Compute the output vector H of the hidden layer

$$net_k = \sum w_{ik} x_i - \theta_k \quad (33)$$
$$H_k = f(net_k) \quad (34)$$

   (b) Compute the output vector Y of the output layer

$$net_j = \sum w_{kj} H_i - \theta_j \quad (35)$$

$$Y_j = f(net_j) \qquad (36)$$

## 4. EMPIRICAL ILLUSTRATION
### 4.1 PURPOSE OF STUDY

The objective of this study is to classify the enterprise distress on the financial structure, debt paying ability, operating ability and earning ability. We furthermore make the comparison of classification performance between the SVM model and BPN model.

### 4.2 DATA SETS

The data of this study were collected from a securities firm's data base in Taiwan. 20 experimental samples, 25 samples for testing data are random select from database.

### 4.3 ENTERPRISE FINANCIALSTATUS INDICATOR

In this study, using financial structure, debt paying ability, operating ability and earning ability are measured attribute. There have 15 measure indexes (as in Table 1). Table 1 is showed as enterprise financial distress index.

### 4.4 PROCESS

Step 1: Data preprocess and variable selection

In this study, the measured attribute are financial structure, debt paying ability, operating ability and earning ability. Table 1 is showed as enterprise financial distress index. From Table 1, the measured attribute's valves of each listed company are $(X_1*7 + X_2*6 + X_3*12)/25$ in financial structure, $(X_4*8 + X_5*8 + X_6*5 + X_7*4)/25$ in Earning ability, $(X_8*7 + X_9*6 + X_{10}*5 + X_{11}*4 + X_{12}*3 /25$ in operating ability, $(X_{13}*8 + X_{14}*7 + X_{15}*10)/25$ in debt paying ability. Table 2 is denoted as the grade of measured attributes (pretreatment training data).

Table 1. Enterprise financial status indicator

| Measured attribute | Financial distress index | Score |
|---|---|---|
| Financial structure | $X_1$:Debt ratio | 7 |
| | $X_2$: Stockholder / total assets | 6 |
| | $X_3$:EPS (Earning per share) | 12 |
| Earning ability | $X_4$:ROA (Return on total assets) | 8 |
| | $X_5$:ROE (Return on Stockholders equity) | 8 |
| | $X_6$:Profit margin | 5 |
| | $X_7$:Earning per share | 4 |
| Operating ability | $X_8$: Fixed Assets turnover ratio | 7 |
| | $X_9$: Account receivable turnover ration | 6 |
| | $X_{10}$: Average collection period | 5 |
| | $X_{11}$ :Inventory turnover ratio | 4 |

|  | $X_{12}$: Average days to sell the Inventory | 3 |
|---|---|---|
| Debt paying ability | $X_{13}$: Current ratio | 8 |
|  | $X_{14}$: Quick ratio | 7 |
|  | $X_{15}$: Asset-liability current ratio | 10 |

Table 2. The grade of measured attributes ( pretreatment training data )

| Listed company | Financial structure | Earning ability | Operating ability | Debt paying ability | attribute of y |
|---|---|---|---|---|---|
| 1 | 0.23 | 0.20 | 0.09 | 0.20 | -1 |
| 2 | 0.18 | 0.18 | 0.10 | 0.21 | -1 |
| 3 | 0.16 | 0.18 | 0.08 | 0.17 | -1 |
| 4 | 0.19 | 0.11 | 0.12 | 0.18 | -1 |
| 5 | 0.20 | 0.22 | 0.11 | 0.19 | -1 |
| 6 | 0.24 | 0.2 | 0.09 | 0.20 | -1 |
| 7 | 0.23 | 0.14 | 0.06 | 0.20 | -1 |
| 8 | 0.20 | 0.08 | 0.07 | 0.10 | 1 |
| 9 | 0.18 | 0.09 | 0.05 | 0.18 | 1 |
| 10 | 0.19 | 0.12 | 0.03 | 0.12 | 1 |
| 11 | 0.22 | 0.13 | 0.04 | 0.15 | -1 |
| 12 | 0.16 | 0.10 | 0.07 | 0.14 | 1 |
| 13 | 0.19 | 0.09 | 0.11 | 0.12 | 1 |
| 14 | 0.15 | 0.18 | 0.16 | 0.10 | -1 |
| 15 | 0.18 | 0.20 | 0.20 | 0.08 | -1 |
| 16 | 0.12 | 0.17 | 0.18 | 0.13 | -1 |
| 17 | 0.21 | 0.18 | 0.10 | 0.12 | -1 |
| 18 | 0.19 | 0.18 | 0.12 | 0.09 | -1 |
| 19 | 0.22 | 0.19 | 0.09 | 0.14 | -1 |
| 20 | 0.20 | 0.15 | 0.15 | 0.09 | -1 |

Step 2: Sample data Processing

In this study, we use 15 financial evaluation indexes, and select 20 training sample data, 25 testing sample data from listed company.  The training experiments were conducted on a small data set.  According to Table 1, we grade of measured attributes. The measured attributes are financial structure, earning ability, operating ability, and debt paying ability; (y), y is sample data decision attribute.  If listed company is special treatment (ST) then y is 1, and otherwise y is -1

Step 3: Solve the Enterprise Financial Distress evaluation problem

 (1) SVM method

In LIBSVM software, we use 3-fold cross-validation to find the best parameter C and γ [3]. The results confirmed that the classification precision of the SVM with radial function (RBF) kernel function was as high as 100% when γ and C where 0.25 and 1.  Then we use the best parameter C and γ to train the whole training set, we have 9 support vector index sets.

The outputs from LIBSVM software are:

    Accuracy = 100% (20/20) (classification)

    Mean square error = 0

    Squared correlation coefficient = 1 (regression)

    Iteration = 108

(2) BPN method

The three parameters of learning rate, momentum and the number of nodes in the hidden layer should be defined for back-propagation network modeling. In the training model, the four factors where fed as input nodes. In this network, each unit of the output layer stands for presence (+1) or absence (-1) of the detected molecule. > 0 as a target value of the presence and < 0 as a target value of the absence. We adopted the range of 0.6-0.9 and 0.1-0.4 to be the decisions of learning rate and momentum [17].

In this illustration, there have four measured attributes; we selected four input node, four hidden nodes, and one output node, that is a good MLP model in this illustration. A very rough rule-of-thumb for number of hidden nodes defined as $h/(5\times(m+n))$, where h, m, and n represent the number of training patterns, output nodes and input nodes respectively. Then the root-mean-square error (RMSE) and classification rate are the measurement indicators to validate the performance of the training model. Through several trail-and-error experiments, the structure of 4 -4 -1 model had the best performance.

The outputs from Mathlib software are:

    Accuracy = 95% (19/20) (classification)

    Iteration = 258

Step 5: Forecast of testing sample data.

We compare the accuracy of different approaches by introducing two error types. Type Ⅰ error refer to the situation when matched data is classified as unmatched one, and Type Ⅱ error refer to unmatched data is classified into matched data. The 25 testing sample data from listed company is denoted as Table 3. The prediction result is listed in Table 4. SVM method shows the best overall prediction accuracy level at 100 % (see Table 4). Using the unmatched and unbalanced testing data, MLP method, shows the best overall prediction accuracy level at 96 % (see Table 5).

Table 4. Result comparison with SVM and MLP with Type Ⅰ and Type Ⅱ error

| Method | Number of sample | Type Ⅰ error | Type Ⅱ error | Error | Accuracy |
|---|---|---|---|---|---|

International Journal of Artificial Intelligence & Applications (IJAIA), Vol.1, No.3, July 2010

| SVM | 25 | 0/25 (0.0%) | 0/25 (0.0%) | 0/25 (0.0%) | 100% |
|---|---|---|---|---|---|
| BPN | 25 | 0/25 (0.0%) | 1/25(4%) | 1/25 (4%) | 96% |

## 5. CONCLUSION

This study constructed an enterprise financial evaluation model based on support vector machine and Back Propagation neural.  First, we define the SVM problem and propose to utilize Lagrange function, KKT condition and Kernel function creates SVM dual form and discriminate function.   Second, we simply define the BPN method and two category model of Multilayer Perception.   Third, a proposed for enterprise distress evaluation is discussed by using SVM and BPN.   Finally, an example result proves the validity of our proposed.   We find SVM method is better than MLP method.

Table 5. The grade of measured attributes (pretreatment testing data )

| Listed company | Financial structure | Earning ability | Operating ability | Debt paying ability | Attribute of y | Correction of attribute y |
|---|---|---|---|---|---|---|
| 1 | 0.22 | 0.20 | 0.10 | 0.20 | -1 | 1 (error) |
| 2 | 0.20 | 0.18 | 0.10 | 0.21 | -1 | 1 (error) |
| 3 | 0.16 | 0.20 | 0.08 | 0.17 | -1 | |
| 4 | 0.19 | 0.15 | 0.12 | 0.18 | -1 | |
| 5 | 0.20 | 0.22 | 0.20 | 0.19 | -1 | |
| 6 | 0.22 | 0.2 | 0.09 | 0.20 | -1 | |
| 7 | 0.23 | 0.18 | 0.06 | 0.20 | -1 | |
| 8 | 0.20 | 0.08 | 0.09 | 0.10 | 1 | |
| 9 | 0.18 | 0.12 | 0.05 | 0.11 | 1 | -1 (error) |
| 10 | 0.19 | 0.15 | 0.03 | 0.12 | 1 | |
| 11 | 0.22 | 0.13 | 0.08 | 0.15 | -1 | |
| 12 | 0.18 | 0.10 | 0.07 | 0.14 | 1 | |
| 13 | 0.19 | 0.09 | 0.12 | 0.12 | 1 | |
| 14 | 0.16 | 0.18 | 0.16 | 0.10 | -1 | |
| 15 | 0.18 | 0.22 | 0.20 | 0.08 | -1 | |
| 16 | 0.12 | 0.17 | 0.18 | 0.15 | -1 | |
| 17 | 0.21 | 0.18 | 0.12 | 0.12 | 1 | |
| 18 | 0.19 | 0.20 | 0.12 | 0.10 | -1 | |
| 19 | 0.22 | 0.15 | 0.09 | 0.14 | -1 | |
| 20 | 0.21 | 0.15 | 0.18 | 0.10 | -1 | 1 (error) |
| 21 | 0.20 | 0.13 | 0.09 | 0.15 | -1 | |
| 22 | 0.18 | 0.15 | 0.07 | 0.18 | 1 | |
| 23 | 0.16 | 0.10 | 0.12 | 0.14 | 1 | |
| 24 | 0.19 | 0.18 | 0.16 | 0.18 | -1 | 1 (error) |
| 25 | 0.16 | 0.20 | 0.20 | 0.09 | -1 | |

**ACKNOWLEDGEMENTS**

I would like to thank the anonymous reviewers for their constructive comments on this paper.

**Ming-Chang Lee** is Assistant Professor of Department of Information Management at Fooyin University and National Kaohsiung University of Applied Sciences. His qualifications include a Master degree in applied Mathematics from National Tsing Hua University and a PhD degree in Industrial Management from Nation Cheng Kung University. His research interests include knowledge management, parallel computing, and data analysis. His publications include articles in the journal of Computer & Mathematics with Applications, International Journal of Operation Research, Computers & Engineering, American Journal of Applied Science and Computers, Industrial Engineering, International Journal innovation and Learning, Int. J. Services and Standards, Lecture Notes in computer Science (LNCS), International Journal of Computer Science and Network Security, Journal of Convergence Information Technology and International Journal of Advancements in computing Technology.

**To Chang** is an Assistant Professor and the Chairman of Department of Information Management at Shu-Te University, Taiwan. His qualifications include Master degree in Computer Science from Naval Postgraduate School, USA and PhD degree in Electronic Engineering from Chung Cheng Institute of Technology, Taiwan. His research interests include Information Security, Management Information Systems, and Enterprise Resource Planning. His publications include articles in International Journal Innovation and Learning, Int. J. Services and Standards, and International Journal of Computer Science and Network Security.

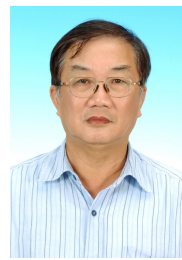

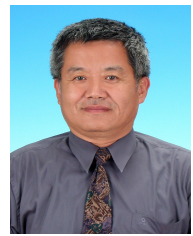